\documentclass[10pt,twocolumn,letterpaper]{article}

\usepackage{multirow}
\usepackage[pagenumbers]{cvpr} 

\definecolor{cvprblue}{rgb}{0.21,0.49,0.74}
\usepackage[pagebackref,breaklinks,colorlinks,allcolors=cvprblue]{hyperref}
\usepackage{xcolor}

\def\ourmodel{PGP-DiffSR\xspace}

\title{PGP-DiffSR: Phase-Guided Progressive Pruning for Efficient Diffusion-based Image Super-Resolution}

\author{Zhongbao Yang \quad Jiangxin Dong \quad Yazhou Yao  \quad Jinhui Tang \quad Jinshan Pan \\
School of Computer Science and Engineering, Nanjing University of Science and Technology\\
{\tt \url{https://github.com/yzb1997/PGP-DiffSR}}
}

\begin{document}
\maketitle

\begin{abstract}
Although diffusion-based models have achieved impressive results in image super-resolution, they often rely on large-scale backbones such as Stable Diffusion XL (SDXL) and Diffusion Transformers (DiT), which lead to excessive computational and memory costs during training and inference.
To address this issue, we develop a lightweight diffusion method, PGP-DiffSR, by removing redundant information from diffusion models under the guidance of the phase information of inputs for efficient image super-resolution.
We first identify the intra-block redundancy within the diffusion backbone and propose a progressive pruning approach that removes redundant blocks while preserving restoration capability.
We note that the phase information of the restored images produced by the pruned diffusion model is not well estimated.
To solve this problem, we propose a phase-exchange adapter module that explores the phase information of the inputs to guide the pruned diffusion model for better restoration performance.
We formulate the progressive pruning approach and the phase-exchange adapter module into a unified model.
Extensive experiments demonstrate that our method achieves competitive restoration quality while significantly reducing computational load and memory consumption.
\vspace{-6mm}
\end{abstract}

\section{Introduction}
\label{sec:intro}
\begin{figure*}[!h] 
\footnotesize
\centering
    \begin{tabular}{c c c c c c c}
            \multicolumn{3}{c}{\multirow{4}*[57.8pt]{
            \hspace{-3.5mm} \includegraphics[width=0.343\linewidth,height=0.285\linewidth]{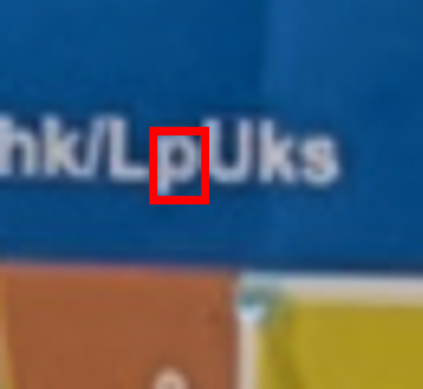}}}
            & \hspace{-4.0mm} \includegraphics[width=0.16\linewidth,height=0.13\linewidth]{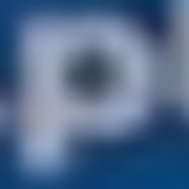} 
            & \hspace{-4.0mm} \includegraphics[width=0.16\linewidth,height=0.13\linewidth]{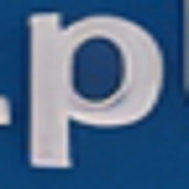}
            & \hspace{-4.0mm} \includegraphics[width=0.16\linewidth,height=0.13\linewidth]{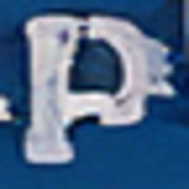} 
            & \hspace{-4.0mm} \includegraphics[width=0.16\linewidth,height=0.13\linewidth]{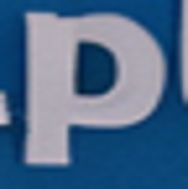} 

              \\

    		\multicolumn{3}{c}{~}                                   
            & \hspace{-4.0mm} (a) LQ patch
            & \hspace{-4.0mm} (b) FaithDiff~\cite{faithdiff}
            & \hspace{-4.0mm} (c) Pruned model of (b) 
            & \hspace{-4.0mm} (d) \ourmodel \\	
            
    	\multicolumn{3}{c}{~} 
            & \hspace{-4.0mm} \includegraphics[width=0.16\linewidth,height=0.13\linewidth]{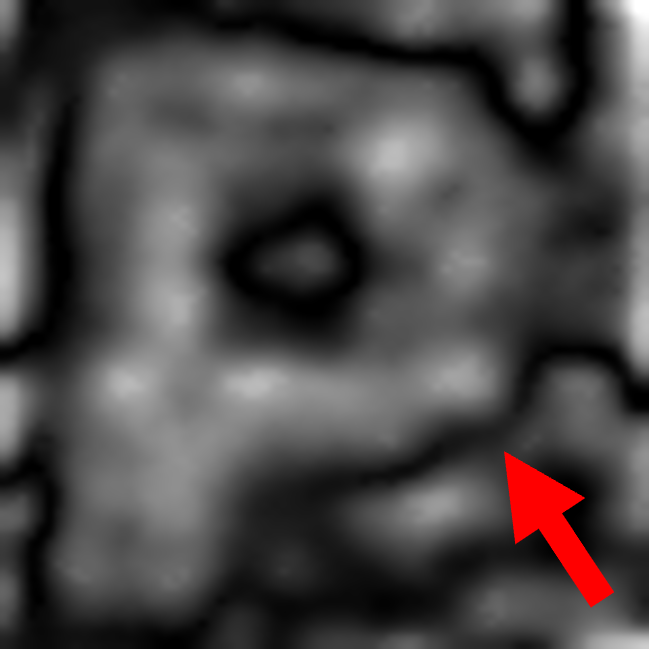} 
            & \hspace{-4.0mm} \includegraphics[width=0.16\linewidth,height=0.13\linewidth]{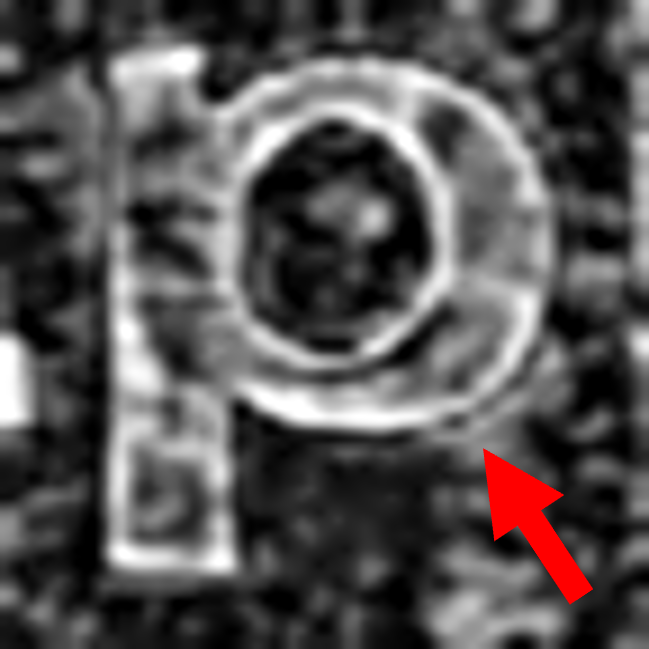}  
            & \hspace{-4.0mm} \includegraphics[width=0.16\linewidth,height=0.13\linewidth]{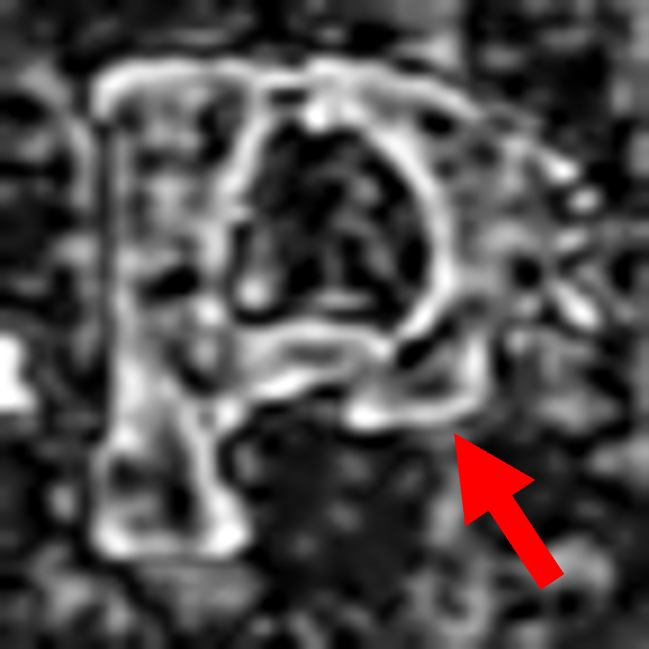}
            & \hspace{-4.0mm} \includegraphics[width=0.16\linewidth,height=0.13\linewidth]{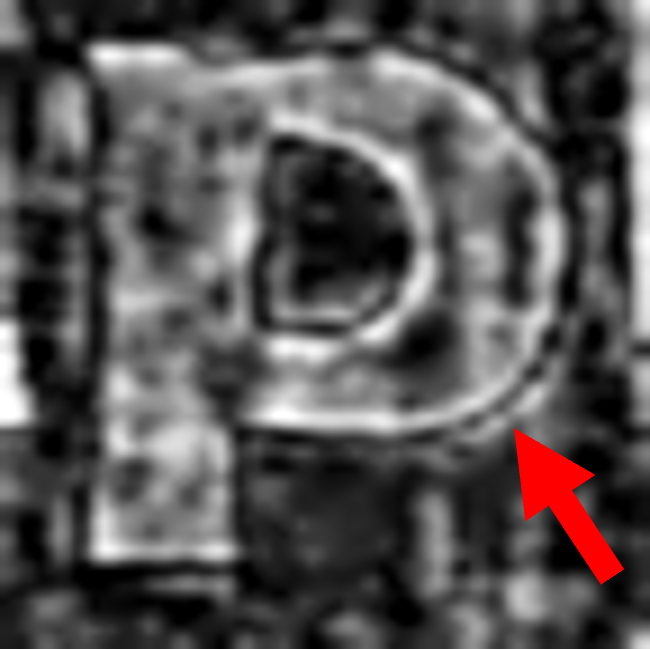}

            \\

    	\multicolumn{3}{c}{\hspace{-4.0mm} LQ Image}
            & \hspace{-4.0mm} (e) POR of (a)
            & \hspace{-4.0mm} (f) POR of (b)  
            & \hspace{-4.0mm} (g) POR of (c)
            & \hspace{-4.0mm} (h) POR of (d)
    \end{tabular}
\vspace{-3mm}
\caption{Effect of the pruning strategy on the image super-resolution. 
(a) denotes the cropped LQ patch from the red box in the input image.
(b) denotes the restored image by FaithDiff~\cite{faithdiff}.
(c) denotes the restored image by applying simple pruning strategy~\cite{adcsr} to FaithDiff~\cite{faithdiff}.
(d) denotes the restored image by the proposed PGP-DiffSR.
(e–h) denote phase-only reconstructions (POR) of (a–d), where POR refers to images reconstructed by inverse Fourier transform after retaining only the Fourier phase.
The structures of restored images by the pruned model are distorted significantly compared to those of LQ image as shown in (e) and (g). 
Therefore, we explore useful structures of LQ images in the frequency domain to facilitate the pruned diffusion model for better image super-resolution. 
}
\vspace{-5mm}
\label{fig1}
\end{figure*}

Image super-resolution (SR) aims to recover high-quality (HQ) images from degraded low-quality (LQ) inputs. 
Recently, the remarkable success of diffusion models has attracted widespread attention in this field~\cite{SUPIR}. 
However, diffusion models (\emph{e.g.}, Stable Diffusion~\cite{sd} and Diffusion Transformer~\cite{sd3}) impose heavy computational demands during training and inference, severely restricting their practicality in resource-constrained environments.
Therefore, there is a great need to develop an efficient diffusion model for image super-resolution.
Existing efficient diffusion methods focus on reducing the training resources, improving the inference efficiency, and lowering the memory usage of diffusion models. 
However, those methods fail to achieve a good trade-off among these three aspects.
To reduce the training resources of the diffusion model, existing control-based diffusion models~\cite{DiffBIR, PASD, SUPIR, dit4sr} use LQ images as conditional signals to control the frozen diffusion backbone and guide image restoration.
Although this strategy significantly reduces training resources, the inference process still relies on the full diffusion model, leading to slow generation speed and high memory consumption.
To further improve inference efficiency, several distillation-based methods~\cite{snapfusion, ticket} transfer knowledge from a large teacher model to a compact student model, thereby reducing memory usage and running time during inference.
Although these methods still depend on a high-capacity teacher network during training, resulting in considerable computational costs.
Afterwards, several training-free acceleration methods~\cite{DPM-Solve, DDIM, deepcache} have been proposed to improve the inference speed without incurring additional training costs.
However, their reliance on modified sampling strategies does not reduce overall resource consumption.
Thus, it remains crucial to explore strategies for achieving a practical balance between training resources, inference efficiency, and memory usage.
A natural solution to reduce computational and memory usage is to prune redundant blocks in the denoising UNet of diffusion models. 
However, the existing method~\cite{fasterdiffusion} only skips the redundant encoder module of the denoising UNet in diffusion models at selected contiguous timesteps during the inference stage, which does not effectively reduce computational costs and memory consumption in the training stage.
In addition, simply applying a pruning strategy~\cite{adcsr} affects the main structures of the restored images (see Figure~\ref{fig1}(c)). 
As the main structures of restored images should be consistent with those of LQ images, and since the phase information usually models these main structures. 
This motivates us to progressively prune the diffusion model and explore the phase information of LQ images to correct the distorted structures generated by the pruned diffusion models for efficient and effective image super-resolution. 

In this paper, we propose \ourmodel, a simple yet effective diffusion-based image super-resolution approach that is efficient during both training and inference.
In contrast to existing methods that only focus on the redundancy of the encoder module of the denoising UNet and lack a systematic analysis of the bottleneck and decoder redundancy, which results in limited pruning gains, we propose a progressive pruning approach to progressively reduce the redundant features from the encoder, bottleneck, and decoder modules in the denoising UNet. 
As the pruning method inevitably affects the restoration of structures and details, we develop a phase-exchange adapter module  to explore the phase information of LQ inputs for better image restoration.
We integrate the proposed progressive pruning approach and phase-exchange adapter module into a lightweight diffusion-based image super-resolution model. We show that the proposed model, i.e., \ourmodel, can be trained on a single RTX 4090 GPU and achieves twice the inference speed compared with existing multi-step diffusion-based methods. 
To further enhance inference efficiency, we develop a one-step efficient diffusion model (named PGP-DiffSR-S1) for image super-resolution based on our PGP-DiffSR model. 
Extensive experimental results demonstrate that the proposed model requires less GPU memory, is more efficient, and achieves favorable performance compared to state-of-the-art methods.
The main contributions are summarized as follows:
\begin{itemize}
    \item We propose a progressive pruning approach to reduce redundant features from the encoder, bottleneck, and decoder in the denoising UNet. 
    
    \item We develop a phase-exchange adapter module to explore the phase information so that the structures and details of latent features can be better preserved for improved image restoration.
    
    \item We further develop an efficient one-step model based on the proposed PGP-DiffSR, which is much more efficient while maintaining competitive performance on super-resolution benchmarks.
\end{itemize}

\vspace{-2mm}
\section{Related Work}
\label{sec:related}

\begin{figure*}[ht] 
    
\centering
\includegraphics[width=1\textwidth]{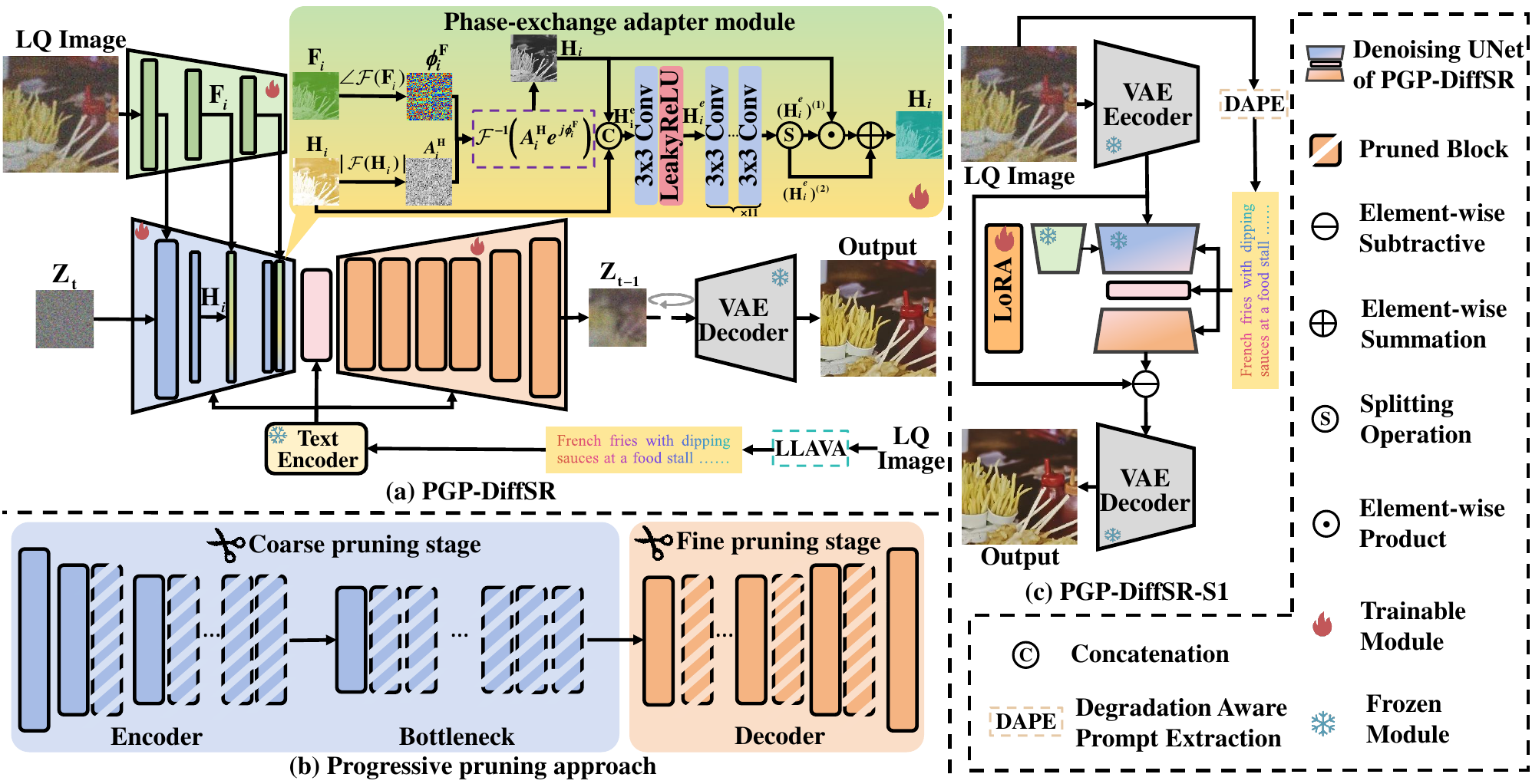}
\vspace{-6mm}
\caption{
Network architectures. (a) The proposed PGP-DiffSR mainly contains a progressive pruning approach i.e., (b), to reduce redundant blocks and a phase-exchange adapter module to explore useful structures of LQ images for better image restoration. 
(c) denotes the one-step version of the proposed PGP-DiffSR.
}
\vspace{-6mm}
\label{fig2}
 
\end{figure*}

\paragraph{Image super-resolution.}
CNN-based methods \cite{smfanet, swinir} are efficient and stable; however, the reconstruction results are often overly smooth due to the convolution operations.
Although GAN-based methods \cite{esrgan, realesrgan, LDL} enhance perceptual realism for the LQ image, they introduce training instability and the risk of mode collapse. 
Recently, diffusion-based image restoration methods~\cite{faithdiff, DiffBIR, SUPIR, PASD, dit4sr} have used strong generative priors to deliver competitive SR performance with a favorable perceptual–fidelity trade-off.
However, their reliance on large-scale backbones (e.g., SD, SDXL, SD3) incurs heavy computing and memory costs, limiting deployment on resource-constrained devices.
\vspace{-3mm}

\paragraph{Efficient diffusion models.}
To reduce the cost of diffusion-based SR, control-based methods~\cite{SUPIR, DiffBIR, PASD, dit4sr} utilize LQ images as conditional signals to guide the frozen diffusion backbone for image restoration, thereby lowering overhead in the training stage.

However, during the inference stage, those methods still require the complete diffusion backbone (many sampling steps with large intermediate activations), resulting in slow runtimes and high memory usage, which leads to slow running speeds and high memory usage on GPUs.
To alleviate the computational burden of diffusion-based models during the inference stage, distillation-based diffusion model methods~\cite{snapfusion, ticket} transfer knowledge from large-scale diffusion models to compact student models, thereby training less complex diffusion models. 
Existing methods for redundancy analysis and pruning of internal modules in diffusion models can reduce the number of parameters and GPU memory usage during the training and inference stages.
However, pruned diffusion models often undermine the prior information of diffusion models, thereby degrading recovery quality. Even with fine-tuning, it still fails to match the performance of the unpruned diffusion model.
Different from existing methods, while removing redundant modules to reduce the computational and memory overhead of the diffusion model, we utilize the phase information of LQ images to guide the pruned diffusion model, thereby enabling it to better recover its original image restoration capability during the fine-tuning process.
\vspace{-3mm}

\section{Proposed Method}
\label{sec:methodology}
\vspace{-1mm}
In this section, we develop an efficient diffusion-based SR method, PGP-DiffSR. 
The proposed method mainly consists of a simple yet effective progressive pruning approach (PPA) and a phase-exchange adapter module (PEAM). The former removes redundant blocks to reduce the computational and memory overhead of the diffusion model, while the latter extracts more informative features for better image restoration.
Figure~\ref{fig2} shows an overview of our method.

\subsection{Progressive pruning approach}
\begin{figure*}[ht] 
\centering
\includegraphics[width=1\textwidth]{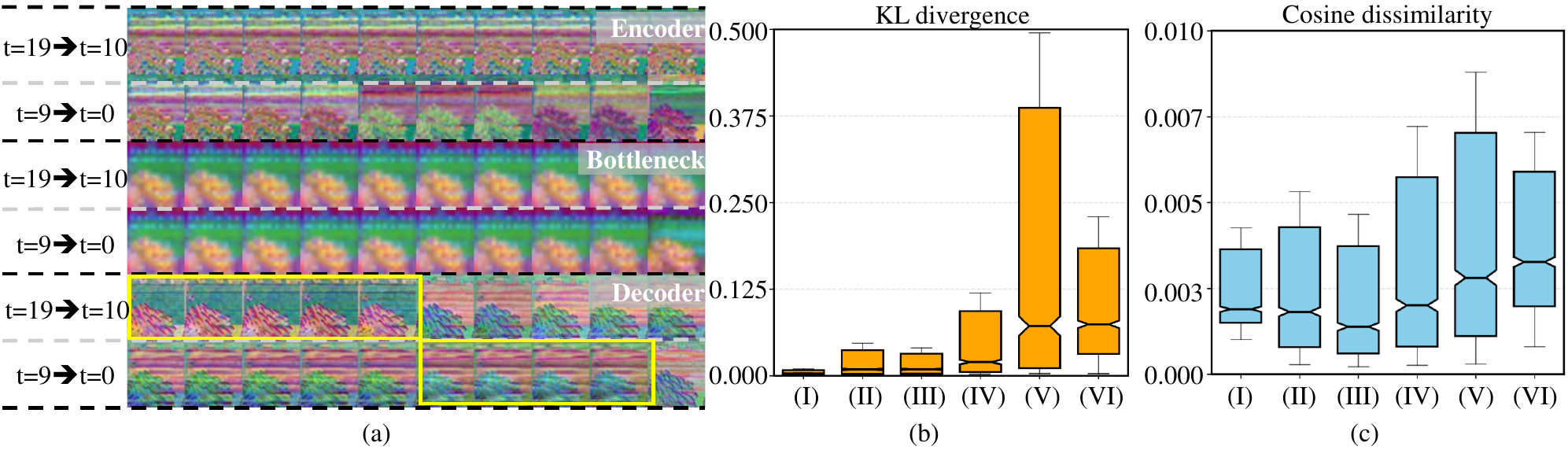}
\vspace{-5mm}
\caption{\textbf{Redundancy analysis in diffusion models.}
  (a) Visualizations of the features in the denoising UNet of diffusion model at each timestep (t=0$\to$t=19).
  (b) Box-plots of KL divergence of the features across adjacent timesteps in the denoising UNet.
  (I)–(III) indicate
  $\{\Psi_{\mathrm{KL}}(\mathbf{E}^{(t+1)}_{s},\mathbf{E}^{(t)}_{s})\}_{s=1,2,4}$;
  (IV) indicates $\Psi_{\mathrm{KL}}(\mathbf{B}^{(t+1)}_{4},\mathbf{B}^{(t)}_{4})$;
  (V) and (VI) indicate
  $\Psi_{\mathrm{KL}}(\mathbf{D}^{(t+1)}_{4},\mathbf{D}^{(t)}_{4})$ and 
  $\Psi_{\mathrm{KL}}(\mathbf{D}^{(t+1)}_{2},\mathbf{D}^{(t)}_{2})$, respectively.
  (c) Box-plots of cosine dissimilarity of the features across adjacent timesteps in the the denoising UNet:
  (I)–(III) indicate
  $\{\Phi_{\mathrm{cos}}(\mathbf{E}^{(t+1)}_{s},\mathbf{E}^{(t)}_{s})\}_{s=1,2,4}$;
  (IV) indicates $\Phi_{\mathrm{cos}}(\mathbf{B}^{(t+1)}_{4},\mathbf{B}^{(t)}_{4})$;
  (V) and (VI) indicate
  $\Phi_{\mathrm{cos}}(\mathbf{D}^{(t+1)}_{4},\mathbf{D}^{(t)}_{4})$ and
  $\Phi_{\mathrm{cos}}(\mathbf{D}^{(t+1)}_{2},\mathbf{D}^{(t)}_{2})$, respectively.
  The yellow boxes in (a) illustrate that feature variations in the decoder of the denoising UNet remain minimal across certain consecutive timesteps, indicating the presence of redundant blocks within the decoder module.
  }
\vspace{-5mm}
\label{fig3}
\end{figure*}

Existing methods~\cite{fasterdiffusion,ticket} achieve efficiency gains in diffusion denoising UNet by reducing the redundancy of the encoder module across consecutive time steps, as this redundancy contributes little to quality but incurs substantial computational costs.
However, those methods reduce redundancy only in the encoder module of the denoising UNet and leave the bottleneck and decoder modules of the denoising UNet insufficiently explored.
We observe redundancy in the bottleneck and decoder modules across some consecutive timesteps (Figure~\ref{fig3}(a)), revealing that these modules contain redundant blocks that offer limited benefits to quality while inflating computational costs.
Therefore, we propose a PPA that consists of a coarse pruning stage and a fine pruning stage to reduce redundant blocks in the encoder, bottleneck, and decoder modules of the denoising UNet.
\vspace{-4mm}

%
{\flushleft \textbf{Coarse pruning stage.}} 
To quantitatively evaluate redundancy blocks in the encoder, bottleneck, and decoder modules of the denoising UNet, we employ cosine-based dissimilarity and Kullback–Leibler (KL) divergence to measure the similarity between features.
First, assuming that the input noise latent $Z_t$ has a spatial resolution of $H \times W$ pixels from the denoising UNet of the diffusion model at timestep $t$, and the spatial resolution of the feature $\mathbf{E}^{(t)}_{s}$ from the encoder module of the denoising UNet at timestep $t$ is $H/s \times W/s$ (where $s$ denotes the downsampling scale factor), we measure the similarity of $\mathbf{E}^{(t)}_{s}$ and $\mathbf{E}^{(t+1)}_{s}$:  
\begin{equation}
\Phi_{\mathrm{cos}}(\mathbf{E}^{(t)}_{s},\mathbf{E}^{(t+1)}_{s})
=
\mathbb{E}\!\left[
1-
\frac{\langle {\mathbf{E}^{(t)}_{s}},\,\mathbf{E}^{(t+1)}_{s}\rangle}
{\|\mathbf{E}^{(t)}_{s}\|_2\,\|\mathbf{E}^{(t+1)}_{s}\|_2}
\right],
\label{eq:E_cos_general}
\end{equation}
\begin{equation}
\Psi_{\mathrm{KL}}(\mathbf{E}^{(t)}_{s},\mathbf{E}^{(t+1)}_{s})
=
\mathbb{E}\!\left[
D_\mathrm{KL}\!\big(\mathcal{S}(\mathbf{E}^{(t)}_{s})\,\|\,\mathcal{S}(\mathbf{E}^{(t+1)}_{s})\big)
\right],
\label{eq:E_kl_general}
\end{equation}
where $D_\mathrm{KL}(\|\cdot\|)$ denotes the KL divergence, $\mathcal{S}(\cdot)$ denotes the softmax operation, $t\in\{0, \dots, T-1, \}$, $s\in\{1,2,4\}$, and $T$ denotes the total number of timesteps in the diffusion process.
Similarly, for the features $\mathbf{B}^{(t+1)}_{s}$ and $\mathbf{B}^{(t)}_{s}$ from the bottleneck module, as well as those $\mathbf{D}^{(t+1)}_{s}$ and $\mathbf{D}^{(t)}_{s}$ from the decoder module in the denoising UNet, their similarities can be obtained by $\Phi_{\mathrm{cos}}(\mathbf{B}^{(t+1)}_{s},\mathbf{B}^{(t)}_{s})$, $\Psi_{\mathrm{KL}}(\mathbf{B}^{(t+1)}_{s},\mathbf{B}^{(t)}_{s})$, $\Phi_{\mathrm{cos}}(\mathbf{D}^{(t+1)}_{s},\mathbf{D}^{(t)}_{s})$, and $\Psi_{\mathrm{KL}}(\mathbf{D}^{(t+1)}_{s},\mathbf{D}^{(t)}_{s})$.
We show the visualization of the features from the denoising UNet and the similarity values of these features in Figure~\ref{fig3}.
In Figure~\ref{fig3}(a), the features of the encoder and bottleneck modules change slightly across adjacent timesteps.
This visual observation is further validated by the statistical results in Figure~\ref{fig3}(b) and (c), which reveal that the encoder and bottleneck modules exhibit consistently low KL divergence and cosine dissimilarity.
Collectively, these results indicate that the encoder and bottleneck modules of the denoising UNet contain redundant information.
Guided by these observations, we preserve the overall architecture of the denoising UNet and prune the encoder and bottleneck modules to a single block.
The effectiveness and efficiency of the coarse pruning stage are demonstrated in Section~\ref{sec:ablation}.

{\flushleft \textbf{Fine pruning stage.}}
Although the features of the decoder module of the denoising UNet exhibit large variations across adjacent timesteps overall (Figure~\ref{fig3}(b) and (c)), we observe that over certain contiguous timesteps, its features change slightly (see the yellow boxes in Figure~\ref{fig3}(a)), indicating that the decoder of the denoising UNet has redundant blocks.
Since relying solely on adjacent timestep analysis cannot identify which blocks in the decoder module of the denoising UNet are redundant, we introduce a fine pruning stage to remove redundant blocks in the decoder module of the denoising UNet under the same timestep conditions.
Let $\mathbf{D}_s^{(In)}$ and $\mathbf{D}_s^{(Out)}$ denote the input and output features from one block of the decoder module in the denoising UNet, respectively. $\Phi_{\mathrm{cos}}(\mathbf{D}_s^{(In)},\mathbf{D}_s^{(Out)})$ and $\Psi_{\mathrm{KL}}(\mathbf{D}_s^{(In)},\mathbf{D}_s^{(Out)})$ are utilized to analyze the redundancy of the block in the decoder module.
We observe that most blocks in the decoder module of the denoising UNet exhibit low similarity (suggesting minimal redundancy), while a subset maintains high similarity, signifying their redundancy in the decoder module.\footnote{The detailed analysis of this observation is included in the supplemental material due to the page limit. We prune blocks from the transformer modules in the encoder, bottleneck, and decoder modules of the denoising UNet during both the coarse and fine pruning stages.}
Thus, we strategically prune these redundant blocks in the decoder module.
Specifically, the block in the decoder module is pruned if it satisfies $\Phi_{\mathrm{cos}}(\mathbf{D}_s^{(In)},\mathbf{D}_s^{(Out)}) \ge 0.2$ and $\Psi_{\mathrm{KL}}(\mathbf{D}_s^{(In)},\mathbf{D}_s^{(Out)}) \ge 1.0$.
Building upon the coarse pruning stage, the fine pruning stage further reduces the complexity (e.g., FLOPs, parameters) of the denoising UNet in the diffusion model.
The proposed PPA effectively reduces memory consumption and inference time in both the training stage and the inference stage for the diffusion model. 
We will show the effectiveness and efficiency of the proposed fine pruning stage and PPA in Section~\ref{sec:ablation} and the supplemental material.
%

\subsection{Phase-exchange adapter module}
Although the PPA can effectively reduce the model parameters and computational cost of diffusion models, it affects the quality of the restored images to some extent, as shown in Figure~\ref{fig1}(c).
For image restoration, the structures of the restored image should be consistent with the structures of the degraded image. However, due to information loss caused by model pruning, the structures of restored images cannot be effectively restored. 
We note that phase information can better distinguish image structures from details, and the features from diffusion models contain rich high-frequency information due to the strong generative ability of diffusion models. 
Therefore, we explore phase information of LQ images and develop a simple yet effective phase-exchange adapter module to facilitate the restoration of image structures in the pruned diffusion models.
Specifically, given the features $\mathbf{F}_i$ that are extracted from the LQ images and $\mathbf{H}_i$ from the encoder module of the denoising UNet at the corresponding levels. 
To transfer the main structures of LQ images to $\mathbf{H}_i$, we extract the phase component $\phi^\mathbf{F}_i$ from $\mathbf{F}_i$ and the amplitude component $A^\mathbf{H}_i$ from $\mathbf{H}_i$ by:
\begin{equation}
\begin{split}
\phi^\mathbf{F}_i &= \angle \mathcal{F}(\mathbf{F}_i), \\
A^\mathbf{H}_i &= |\mathcal{F}(\mathbf{H}_i)|,
\end{split}
\end{equation}
where $\mathcal{F}(\cdot)$ denotes the Fourier transform, $\angle(\cdot)$ extracts the phase spectrum, and $|\cdot|$ denotes the magnitude (amplitude) spectrum. 
We then combine the extracted amplitude $A^\mathbf{H}_i$ and phase $\phi^\mathbf{F}_i$ to construct the exchanged feature $\tilde{\mathbf{H}}_i$ by:

\begin{equation}
\tilde{\mathbf{H}}_i = \mathcal{F}^{-1}\!\left(A^\mathbf{H}_ie^{j \phi^\mathbf{F}_i}\right),
\end{equation}
where $\mathcal{F}^{-1}(\cdot)$ denotes the inverse Fourier transform.
Although $\tilde{\mathbf{H}}_i$ contains valuable phase information from LQ images, it inevitably disrupts the intrinsic feature distribution in the diffusion model.
To address this issue, we develop a simple fusion method to treat $\tilde{\mathbf{H}}_i$ as auxiliary phase guidance to adaptively refine the original encoder feature $\mathbf{H}_i$.
Specifically, we concatenate $\tilde{\mathbf{H}}_i$ and $\mathbf{H}_i$ along the channel dimension to obtain an intermediate feature $\mathbf{H}^e_i \in \mathbb{R}^{2C \times H \times W}$. Then, a convolution with a filter size of $3\times3$, followed by a LeakyReLU activation, is applied to generate a refined feature $\tilde{\mathbf{H}}^e_i \in \mathbb{R}^{64 \times H \times W}$. 
We further apply eleven stacked convolutional layers with a filter size of $3\times3$ to generate $\hat{\mathbf{H}}^e_i \in \mathbb{R}^{2C \times H \times W}$. 
The output $\hat{\mathbf{H}}^e_i$ is then split into two components, i.e., ${(\hat{\mathbf{H}}^e_i)}^{(1)}$ and ${(\hat{\mathbf{H}}^e_i)}^{(2)}$,  along the channel dimension. 
Finally, we update $\tilde{H}_i$ by:
\begin{equation}
\hat{\mathbf{H}}_i = {(\hat{\mathbf{H}}^e_i)}^{(1)} \odot \tilde{\mathbf{H}}_i + {(\hat{\mathbf{H}}^e_i)}^{(2)},
\end{equation}
where $\odot$ denotes multiplication in the element. 
Finally, the encoder module of the denoising UNet in the pruned diffusion model takes $\hat{\mathbf{H}}_i$ as the updated features for image restoration.
We formulate the proposed PPA and PEAM into a unified diffusion model for efficient image super-resolution. To further reduce computational costs, we develop a one-step diffusion model based on the commonly used knowledge strategy~\cite{pisasr}. Instead of using pretrained generative diffusion models as baseline models~\cite{pisasr, osediff, tsdsr}, our method adopts the proposed method as the baseline model, which is simple yet effective (see the analysis in Section~\ref{sec:experiments}). 

\section{Experiments}
\label{sec:experiments}
%
In this section, we present the training settings and the implementation details of our method. 
Subsequently, we evaluate our approach using existing public datasets and compare it with other state-of-the-art methods. 
Due to page limitations, additional comparison results are included in the supplemental material.
%

\subsection{Datasets and implementation details}
{\flushleft \textbf{Dataset.}} Similar to existing methods, e.g.,~\cite{faithdiff}, we use the HQ images from LSDIR~\cite{LSDIR}, DIV2K~\cite{DIV2K}, DIV8K~\cite{div8k}, Flickr2K~\cite{Flicker2k}, and $10,000$ face images from FFHQ~\cite{FFHQ} as the training dataset and generate LQ images by the same method as~\cite{PASD}
In addition, we employ LLAVA~\cite{LLAVA} to generate textual descriptions for each image.
We evaluate PGP-DiffSR and PGP-DiffSR-S1 on three commonly used benchmark datasets, including RealSR~\cite{RealSR}, DrealSR~\cite{DrealSR}, and RealPhoto60~\cite{SUPIR}.
\vspace{-2mm}

\begin{table*}[!t]
\scriptsize
\setlength{\tabcolsep}{10.5pt}
\caption{Efficiency comparisons of diffusion-based SR methods.
All metrics are calculated only for the consumption involved in the diffusion process.
All methods are tested on a machine with an A800 GPU. The best and second performances are highlighted in {\color{red}red} and {\color{blue}blue}, respectively. ``Memory'' denotes the GPU memory usage during the model inference stage.}
\centering
\vspace{-3mm}
\begin{tabular}{lcccccccc}
\toprule
                  & DiffBIR~\cite{DiffBIR} & PASD~\cite{PASD}          & SeeSR~\cite{SeeSR} & DreamClear~\cite{dreamclear} & SUPIR~\cite{SUPIR}      & DiT4SR~\cite{dit4sr}    & FaithDiff~\cite{faithdiff}  & Ours                           \\ \midrule
Inference Step    & 50                     & 20                        & 50                 &  50                          & 50                      & 40                      & 20                          & 20                             \\
Running Time [s]  & 7.93                   & 7.31                      & 10.30               &  7.58                        & 11.40                    & 32.90                    & \textcolor{blue}{2.55}      & \textcolor{red}{0.78}         \\ 
Memory [GB]       & 15.40                   & \textcolor{blue}{6.32}    & 13.20               &  19.90                        & 28.80                    & 22.30                    & 10.40                        & \textcolor{red}{5.69}          \\
\bottomrule
\end{tabular}
\vspace{-3mm}
\label{runningtime}
\end{table*}
\begin{table*}[!h]
\scriptsize
\setlength{\tabcolsep}{2.6pt}
\centering
\caption{
Quantitative comparison with state-of-the-art methods on the RealSR~\cite{RealSR}, DrealSR~\cite{DrealSR} and RealPhoto60~\cite{SUPIR} datasets for the PGP-DiffSR.
The best and second performances are marked in \textcolor{red}{red} and \textcolor{blue}{blue}, respectively.
 ``$\downarrow$'' and ``$\uparrow$'' denote that lower and higher values are better, respectively.
}
\vspace{-3mm}
\begin{tabular}{l|l|cccccccccc}
\toprule
    \textbf{Datasets}                   
    & \textbf{Metrics}          & Real-ESRGAN~\cite{realesrgan} & BSRGAN~\cite{BSRGAN}      & StableSR~\cite{StableSR} & DiffBIR~\cite{DiffBIR}     & PASD~\cite{PASD}          & SeeSR~\cite{SeeSR}        & SUPIR~\cite{SUPIR}        & DiT4SR~\cite{dit4sr}      & FaithDiff~\cite{faithdiff} & Ours \\ \midrule

\multirow{6}{*}{\textbf{RealSR}}
    & LPIPS$\downarrow$         & \textcolor{blue}{0.2709}      & \textcolor{red}{0.2656}   & 0.3205                   & 0.3650                     & 0.2879                    & 0.3007                    & 0.3653                    & 0.3166                    & 0.2832                     & 0.3208 \\
    & DISTS$\downarrow$         & \textcolor{blue}{0.2061}      & 0.2124                    & 0.2258                   & 0.2399                     & \textcolor{red}{0.2040}   & 0.2224                    & 0.2439                    & 0.2230                    & 0.2076                     & 0.2296 \\
    & NIQE$\downarrow$          & 5.797                         & 5.644                     & 5.547                    & 5.839                      & 6.018                     & \textcolor{red}{5.400}    & 6.958                     & 6.007                     & \textcolor{blue}{5.457}    & 5.467  \\
    & CLIP-IQA+$\uparrow$       & 0.4323                        & 0.4576                    & 0.4836                   & \textcolor{blue}{0.5970}   & 0.4833                    & 0.5790                    & 0.5063                    & 0.5406                    & 0.5847                     & \textcolor{red}{0.6056} \\
    & MUSIQ$\uparrow$           & 60.37                         & 63.28                     & 65.55                    & 69.28                      & 59.65                     & \textcolor{blue}{69.82}   & 59.30                     & 67.91                     & 68.96                      & \textcolor{red}{70.17} \\
    & MANIQA$\uparrow$          & 0.5492                        & 0.5416                    & 0.6194                   & 0.6502                     & 0.5347                    & 0.6445                    & 0.5629                    & 0.6567                    & \textcolor{blue}{0.6793}   & \textcolor{red}{0.6841} \\ \midrule

\multirow{6}{*}{\textbf{DrealSR}}
    & LPIPS$\downarrow$         & \textcolor{red}{0.2819}        & \textcolor{blue}{0.2858} & 0.3552                   & 0.4669                     & 0.3174                    & 0.3174                    & 0.4120                    & 0.3706                     & 0.3459                      & 0.3773 \\
    & DISTS$\downarrow$         & \textcolor{red}{0.2089}        & \textcolor{blue}{0.2144} & 0.2378                   & 0.2882                     & 0.2208                    & 0.2315                    & 0.2721                    & 0.2453                     & 0.2386                      & 0.2549 \\
    & NIQE$\downarrow$          & 6.699                          & 6.540                    & 6.398                    & 6.332                      & 7.642                     & 6.405                     & 9.033                     & 6.774                      & \textcolor{blue}{6.184}     & \textcolor{red}{6.171} \\
    & CLIP-IQA+$\uparrow$       & 0.4088                         & 0.4174                   & 0.4319                   & \textcolor{blue}{0.5770}   & 0.4401                    & 0.5428                    & 0.4452                    & 0.5438                     & 0.5649                      & \textcolor{red}{0.6097} \\
    & MUSIQ$\uparrow$           & 54.28                          & 57.16                    & 57.61                    & \textcolor{blue}{66.14}    & 50.28                     & 65.09                     & 53.01                     & 64.21                      & 65.80                       & \textcolor{red}{67.50} \\
    & MANIQA$\uparrow$          & 0.4900                         & 0.4855                   & 0.5516                   & 0.6209                     & 0.4677                    & 0.6043                    & 0.5121                    & 0.6175                     & \textcolor{blue}{0.6386}    & \textcolor{red}{0.6530} \\ \midrule

\multirow{4}{*}{\textbf{RealPhoto60}}
    & NIQE$\downarrow$          & 3.9287                         & 5.3832                   & 4.5180                   & 4.6963                     & 4.2567                    & 4.0193                    & \textcolor{red}{3.1617}   & \textcolor{blue}{3.2428}   & 3.9199                      & 3.5415 \\
    & CLIP-IQA+$\uparrow$       & 0.4395                         & 0.3397                   & 0.3809                   & 0.5410                     & 0.4853                    & 0.5956                    & 0.5508                    & \textcolor{red}{0.6329}    & 0.5761                      & \textcolor{blue}{0.6166} \\
    & MUSIQ$\uparrow$           & 59.28                          & 45.46                    & 53.35                    & 63.53                      & 63.67                     & 71.76                     & 70.19                     & \textcolor{red}{73.41}     & 72.02                       & \textcolor{blue}{72.81} \\
    & MANIQA$\uparrow$          & 0.5079                         & 0.3719                   & 0.4996                   & 0.5881                     & 0.5297                    & 0.6183                    & 0.6189                    & \textcolor{blue}{0.6540}   & \textcolor{red}{0.6600}     & 0.6452 \\ \bottomrule

\end{tabular}
\label{tab1}
\vspace{-3mm}
\end{table*}
\begin{figure*}[!h] 
\footnotesize
\centering
\resizebox{1.015\linewidth}{!}{
    \begin{tabular}{c c c c c c c}
            \multicolumn{3}{c}{\multirow{5}*[57.8pt]{
            \hspace{-4mm} \includegraphics[width=0.325\linewidth,height=0.285\linewidth]{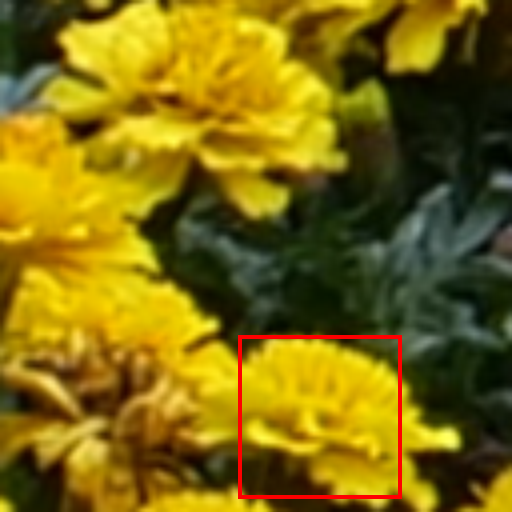}}}
            & \hspace{-4.0mm} \includegraphics[width=0.15\linewidth,height=0.13\linewidth]{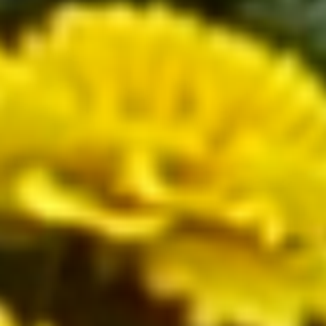}
            & \hspace{-4.0mm} \includegraphics[width=0.15\linewidth,height=0.13\linewidth]{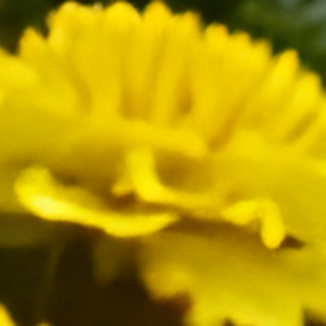} 
            & \hspace{-4.0mm} \includegraphics[width=0.15\linewidth,height=0.13\linewidth]{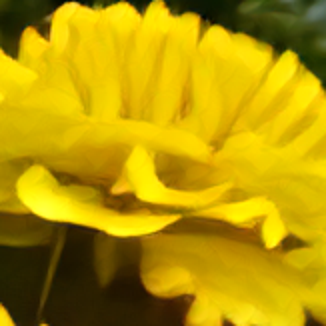} 
            & \hspace{-4.0mm} \includegraphics[width=0.15\linewidth,height=0.13\linewidth]{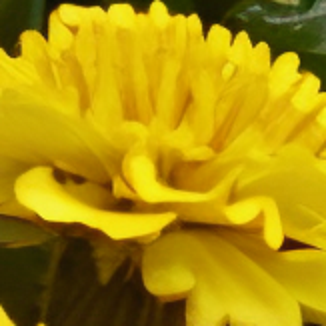} 

              \\

    		\multicolumn{3}{c}{~}                                   
            & \hspace{-4.0mm} (a) LQ patch
            & \hspace{-4.0mm} (b) Real-ESRGAN~\cite{realesrgan} 
            & \hspace{-4.0mm} (c) BSRGAN~\cite{BSRGAN}
            & \hspace{-4.0mm} (d) SeeSR~\cite{SeeSR} \\	
            
    	\multicolumn{3}{c}{~} 
            & \hspace{-4.0mm} \includegraphics[width=0.15\linewidth,height=0.13\linewidth]{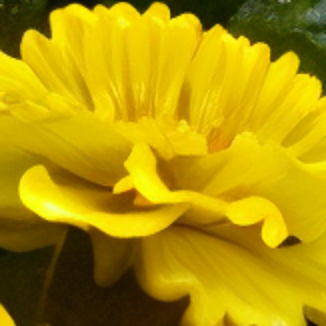}  
            & \hspace{-4.0mm} \includegraphics[width=0.15\linewidth,height=0.13\linewidth]{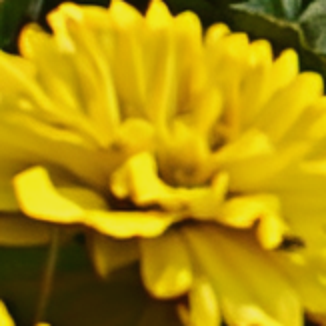}
            & \hspace{-4.0mm} \includegraphics[width=0.15\linewidth,height=0.13\linewidth]{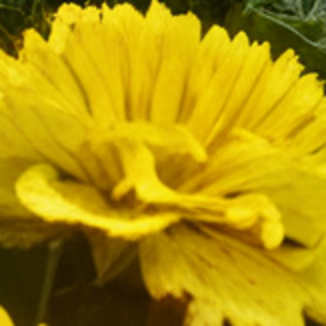} 
            & \hspace{-4.0mm} \includegraphics[width=0.15\linewidth,height=0.13\linewidth]{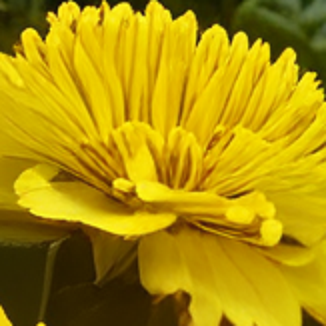} 
            \\

    	\multicolumn{3}{c}{\hspace{-4.0mm} LQ image from DrealSR~\cite{DrealSR}}
            & \hspace{-4.0mm} (e) DiffBIR~\cite{DiffBIR} 
            & \hspace{-4.0mm} (f) DiT4SR~\cite{dit4sr} 
            & \hspace{-4.0mm} (g) FaithDiff~\cite{faithdiff} 
            & \hspace{-4.0mm} (h) Ours

    \end{tabular}
 }
\vspace{-3mm}
\caption{
Image SR results ($\times 2$) on the DrealSR~\cite{DrealSR} dataset for the PGP-DiffSR.
The restored results in (b) to (g) fail to fully restore the richer textures in petal on the flower.
In contrast, our method generates the images with clearer and finer textures.
}
\vspace{-6mm}
\label{fig4}
\end{figure*} 

{\flushleft \textbf{Implementation details.}}
For the proposed PGP-DiffSR, we adopt the diffusion model from~\cite{faithdiff} as the base model.
In the fine pruning stage of PPA, we further set the number of blocks at the smallest-scale in the decoder to 4.
During training, after each pruning stage, we employ an adapter~\cite{T2I_adapter} to extract features from the LQ image and then perform fine-tuning $10{,}000$ iterations.
The initial learning rate for all components is set to $5 \times 10^{-5}$ and is updated according to the cosine annealing schedule~\cite{cosine}.
We employ the AdamW optimizer~\cite{adamw} with its default parameters. 
we set the patch size to $512 \times 512$ patches, and the batch size is set to 128. 
For inference, we adopt the Euler scheduler~\cite{Eulerscheduler} with 20 sampling steps, and the classifier-free guidance~\cite{cfg} scale is set to 5. 
Finally, the proposed PGP-DiffSR model is efficiently trained on one RTX $4090$ GPU.
For PGP-DiffSR-S1, we use PGP-DiffSR as the base model.  
We follow the method in~\cite{SeeSR} and adopt the degradation-aware prompt extraction (DAPE) model to generate text prompts for each image.
The proposed PGP-DiffSR-S1 is trained in two stages. 
In the first stage, the model is trained for $4,000$ iterations using MSE loss and LPIPS~\cite{LPIPS} loss. 
In the second stage, it is further trained with LPIPS loss~\cite{LPIPS} and Classifier Score Distillation loss~\cite{csd} for a total of $14,000$ iterations.
Both training stages use the Adam optimizer~\cite{adamw} with a learning rate of $5 \times 10^{-5}$.
The trainable LoRA modules employ a Gaussian distribution with a rank of 16.
The batch size is set to 16, and the patch size is set to $512 \times 512$.
%

\subsection{Comparisons with the state-of-the-art}
\vspace{-2mm}
We compare our method PGP-DiffSR with the state-of-the-art GAN-based methods, including Real-ESRGAN~\cite{realesrgan} and BSRGAN~\cite{BSRGAN}, as well as diffusion-based methods, including StableSR~\cite{StableSR}, DiffBIR~\cite{DiffBIR}, PASD~\cite{PASD}, SeeSR~\cite{SeeSR}, SUPIR~\cite{SUPIR}, DiT4SR~\cite{dit4sr}, and FaithDiff~\cite{faithdiff}. 
For PGP-DiffSR-S1, we compare it with the state-of-the art one-step diffusion-based image restoration methods, including SinSR~\cite{sinsr}, OSEDiff~\cite{osediff}, AddSR~\cite{addsr}, TVT~\cite{tvt}, PiSA-SR~\cite{pisasr}, and TSD-SR~\cite{tsdsr}.
To comprehensively evaluate both fidelity and perceptual quality, we adopt reference-based metrics (i.e., LPIPS~\cite{LPIPS} and DISTS~\cite{dists}) to quantify fidelity and perceptual quality. 
In addition, we employ non-reference metrics (i.e., NIQE~\cite{niqe}, CLIP-IQA+~\cite{clipiqa}, MUSIQ~\cite{musiq}, and MANIQA~\cite{maniqa}) to further measure the perceptual realism and naturalness of the restored images.
\vspace{-2mm}
%
\begin{table*}[!h]
\scriptsize
\setlength{\tabcolsep}{6.5pt}
\centering
\caption{
Quantitative comparison with state-of-the-art methods on the RealSR~\cite{RealSR}, DrealSR~\cite{DrealSR} and RealPhoto60~\cite{SUPIR} datasets for the PGP-DiffSR-S1.
The best and second performances are marked in \textcolor{red}{red} and \textcolor{blue}{blue}, respectively.
 ``$\downarrow$'' and ``$\uparrow$'' denote that lower and higher values are better, respectively.
}
\vspace{-3mm}
\begin{tabular}{c|l|cccccccc}
\toprule
\textbf{Datasets}                       
& \textbf{Metrics}       &SinSR~\cite{sinsr}        &OSEDiff-S1~\cite{osediff}    &AddSR~\cite{addsr}       & TVT~\cite{tvt}                &PiSA-SR-S1~\cite{pisasr}       &HYPIR~\cite{hypir}             &TSD-SR-SD3~\cite{tsdsr}       & Ours \\ \midrule

\multirow{6}{*}{\textbf{RealSR}}
& LPIPS$\downarrow$      & 0.3189                   & 0.2920                      & 0.3141                  & \textcolor{red}{0.2597}       & \textcolor{blue}{0.2672}      & 0.3046                        & 0.2823                       & 0.3304 \\
& DISTS$\downarrow$      & 0.2347                   & 0.2127                      & 0.2199                  & \textcolor{blue}{0.2061}      & \textcolor{red}{0.2044}       & 0.2238                        & 0.2200                       & 0.2414 \\
& NIQE$\downarrow$       & 6.312                    & 5.636                       & 6.186                   & 5.923                         & 5.506                         & 5.497                         & \textcolor{blue}{5.106}      & \textcolor{red}{5.029} \\
& CLIP-IQA+$\uparrow$    & 0.4401                   & 0.5533                      & 0.4443                  & 0.5480                        & 0.5635                        & 0.5363                        & \textcolor{blue}{0.5707}     & \textcolor{red}{0.5768} \\
& MUSIQ$\uparrow$        & 60.64                    & 69.08                       & 62.66                   & 69.89                         & 70.15                         & 66.42                         & \textcolor{red}{71.26}       & \textcolor{blue}{71.12} \\
& MANIQA$\uparrow$       & 0.5384                   & 0.6331                      & 0.5617                  & 0.6232                        & \textcolor{blue}{0.6552}      & 0.6510                        & 0.6307                       & \textcolor{red}{0.6576} \\ \midrule

\multirow{6}{*}{\textbf{DrealSR}}
& LPIPS$\downarrow$      & 0.3642                   & 0.2968                      & 0.3080                  & \textcolor{red}{0.2900}       & \textcolor{blue}{0.2959}      & 0.3356                        & 0.3109                       & 0.3425 \\
& DISTS$\downarrow$      & 0.2476                   & \textcolor{red}{0.2165}     & 0.2207                  & 0.2205                        & \textcolor{blue}{0.2169}      & 0.2333                        & 0.2210                       & 0.2461 \\
& NIQE$\downarrow$       & 6.954                    & 6.490                       & 7.477                   & 7.033                         & 6.170                         & 6.385                         & \textcolor{blue}{5.880}      & \textcolor{red}{5.849} \\
& CLIP-IQA+$\uparrow$    & 0.4062                   & 0.5181                      & 0.3949                  & 0.5226                        & 0.5290                        & 0.4885                        & \textcolor{blue}{0.5292}     & \textcolor{red}{0.5505} \\
& MUSIQ$\uparrow$        & 55.44                    & 64.65                       & 53.50                   & 65.56                         & 66.11                         & 61.03                         & \textcolor{blue}{66.74}      & \textcolor{red}{67.45} \\
& MANIQA$\uparrow$       & 0.4894                   & 0.5895                      & 0.4760                  & 0.5775                        & \textcolor{blue}{0.6145}      & 0.5993                        & 0.5776                       & \textcolor{red}{0.6205} \\ \midrule

\multirow{4}{*}{\textbf{RealPhoto60}}
& NIQE$\downarrow$       & --                        & 3.7250                     & 4.0383                  & 4.4058                        & \textcolor{blue}{3.5491}      & 3.6104                        & 3.6678                      & \textcolor{red}{3.3447} \\
& CLIP-IQA+$\uparrow$    & --                        & 0.5725                     & 0.4901                  & 0.5700                        & 0.5699                        & \textcolor{blue}{0.5789}      & 0.5310                      & \textcolor{red}{0.6204} \\
& MUSIQ$\uparrow$        & --                        & 70.46                      & 65.77                   & 70.57                         & \textcolor{blue}{72.13}       & 70.97                         & 71.44                       & \textcolor{red}{74.18} \\
& MANIQA$\uparrow$       & --                        & 0.5990                     & 0.5475                  & 0.5943                        & \textcolor{blue}{0.6259}      & 0.6240                        & 0.5573                      & \textcolor{red}{0.6437} \\ \bottomrule

\end{tabular}
\vspace{-5mm}
\label{tab2}
\end{table*}
{\flushleft \textbf{Efficiency comparisons.}}
Benefiting from our PPA, we markedly reduce the computational cost per-step of the denoising UNet while preserving generation quality. 
On a single A800 GPU, we evaluate the running time of diffusion-based SR methods~\cite{DiffBIR, PASD, SeeSR,dreamclear, SUPIR, dit4sr, faithdiff} on image patches with a size of $1024\times1024$ pixels (each method uses its default number of steps). Table~\ref{runningtime} shows that the proposed method has an average time of 0.780s with 20 denoising sample steps, which is significantly faster than other diffusion-based SR methods while achieving the lowest GPU memory usage.
\vspace{-2.5mm}

{\flushleft \textbf{Quantitative comparisons.}}
The quantitative results of PGP-DiffSR are presented in Table~\ref{tab1}.
For the RealSR~\cite{RealSR} and DrealSR~\cite{DrealSR} datasets.
Although existing approaches, GAN-based methods~\cite{realesrgan, BSRGAN} achieve better performance on fidelity metrics such as LPIPS~\cite{LPIPS} and DISTS~\cite{dists}, they suffer from notable limitations in perceptual quality metrics.
Our proposed method outperforms competing methods~\cite{realesrgan, BSRGAN, StableSR, DiffBIR, PASD, SeeSR, SUPIR, dit4sr, faithdiff} in terms of the metrics NIQE~\cite{niqe}, CLIP-IQA+~\cite{clipiqa}, MUSIQ~\cite{musiq}, and MANIQA~\cite{maniqa}.
On the RealPhoto60~\cite{SUPIR} dataset, our method also achieves competitive results. These results demonstrate that our method can still obtain high-quality restoration results with a faster inference speed and lower model complexity.
Table~\ref{tab2} shows the quantitative results of PGP-DiffSR-S1.
Compared with the diffusion-based one-step image restoration method~\cite{sinsr, osediff, addsr, tvt, pisasr, hypir, tsdsr}, our proposed method achieves better results in terms of non-reference metrics (i.e., NIQE~\cite{niqe}, MUSIQ~\cite{musiq}, and MANIQA~\cite{maniqa}.) on three datasets, increasing MUSIQ~\cite{musiq} by at least 2.15 on the RealPhoto60~\cite{SUPIR} dataset.
Our method also achieves competitive results in terms of LPIPS~\cite{LPIPS} and DISTS~\cite{dists} metrics on the RealSR~\cite{RealSR} and DrealSR~\cite{DrealSR} datasets.
\vspace{-2mm}

{\flushleft \textbf{Qualitative comparisons.}}
Figure~\ref{fig4} shows visual comparisons of the proposed PGP-DiffSR and other methods evaluated on the DrealSR~\cite{DrealSR} dataset.
Our pruning-based approach, guided by the phase information of the LQ image, is able to generate high-quality restorations with richer textures and sharper edges while maintaining a low computational cost, as shown in Figure~\ref{fig4}(h).
\vspace{-2mm}

\section{Analysis and discussion}
\label{sec:ablation}
In this section, we conduct a detailed analysis of the proposed method and validate the effectiveness of its main components.
For the ablation studies in this section, we evaluate our method and all the baseline methods on the DrealSR~\cite{DrealSR} dataset.
We train our method and all the baseline methods using a batch size of 128 to illustrate the contributions of each component in our method.
\vspace{-1.5mm}

\subsection{Effect of the proposed PPA}
%
The proposed PPA is employed to reduce the computational complexity (i.e., FLOPs and parameters) of the denoising UNet in the diffusion-based image restoration method.
To demonstrate the effectiveness of the PPA, we compare our proposed method with the baseline full diffusion model (``\text{w/o PPA}'' for short).
Table~\ref{tab4} shows that PPA reduces the computational cost by a significant margin while maintaining image quality comparable to that of the full model. 
In detail, compared with the baseline method, our method achieves a 43\% reduction in FLOPs and a 62\% reduction in the number of parameters, while NIQE decreases by 0.013 and DISTS decreases by 0.0094 (see comparisons of ``$\text{w/o~PPA}$'' and ``Ours'' in Table~\ref{tab4}).
To further examine the effect of the two core stages (i.e., the coarse pruning stage and the fine pruning stage) of PPA, we first compare ``$\text{w/o~PPA}$'' with the baseline method that uses only the coarse pruning stage (``${\text{w/~Coarse}}$'' for short).
Table~\ref{tab4} shows that, compared with the ``w/o PPA,'' pruning only the encoder and bottleneck modules of the denoising UNet already yields substantial savings, as it achieves 30\% fewer FLOPs and 37\% fewer parameters while maintaining comparable performance (see comparisons of ``${\text{w/~Coarse}}$'' and ``${\text{w/o~PPA}}$'' in Table~\ref{tab4}), demonstrating that the coarse pruning stage alone is effective in removing redundant blocks in the encoder and bottleneck modules of the denoising UNet.
Building on ``${\text{w/~Coarse}}$'', we then introduce the fine pruning stage to form our proposed method and compare it with ``${\text{w/~Coarse}}$''. The results show that the fine pruning stage not only further prunes redundant blocks in the decoder module of the denoising UNet but also improves image restoration quality compared with ``${\text{w/~Coarse}}$''. Using the fine pruning stage, it achieves an additional 26\% reduction in FLOPs and a 39\% reduction in parameters, while DISTS drops by 0.0193 and MANIQA increases by 0.0026 (see comparisons of ``Ours'' and ``${\text{w/~Coarse}}$'' in Table~\ref{tab4}).
This illustrates that the fine pruning stage further efficiently removes redundant blocks in the decoder module of the denoising UNet for the pruned diffusion model in the fine pruning stage. 
Together, both stages efficiently prune the denoising UNet.
%

\begin{table}[!t]
\scriptsize
\setlength{\tabcolsep}{7pt}
\centering
\caption{
Quantitative evaluations of the proposed PPA on the DrealSR~\cite{DrealSR} dataset.
 ``$\downarrow$'' and ``$\uparrow$'' denote that lower and higher values are better, respectively.
The FLOPs and model parameters are evaluated on image patches with the size of $1024 \times 1024$ pixels based on the protocols of existing methods with an NVIDIA A800 80GB GPU.
}
\vspace{-3mm}
\begin{tabular}{lcccc}
\toprule
Methods                             & $\text{w/o~PPA}$              & $\text{w/~Coarse}$                         & $\text{w/~Unfine}$                         & Ours             \\
\midrule
NIQE $\downarrow$                   & 6.184                         & 6.8470                                     & 6.4303                                     & \textbf{6.171}   \\
MANIQA$\uparrow$                    & 0.6380                        & 0.6504                                     & 0.6399                                     & \textbf{0.6530}  \\
DISTS$\downarrow$                   & 0.2390                        & 0.2489                                     & 0.2570                                     & \textbf{0.2296}  \\
FLOPs (G)$\downarrow$               & 2993                          & 2106                                       & 2106                                       & \textbf{1555}    \\
Model parameters (M)$\downarrow$    & 2567                          & 1624                                       & 1624                                       & \textbf{984.5}   \\
\bottomrule
\label{tab4}
\end{tabular}
\vspace{-5mm}
\end{table}
\begin{table}[!t]
\centering
\caption{
Quantitative evaluations of the proposed PEAM on the DrealSR~\cite{DrealSR} dataset.
 ``$\downarrow$'' and ``$\uparrow$'' denote that lower and higher values are better, respectively.
}
\vspace{-3mm}
{\scriptsize
\setlength{\tabcolsep}{7.5pt}
\begin{tabular}{lcccc}
\toprule
\textbf{Metrics}                     & w/ Add    & w/ Phase change       & w/ Fusion conv      & Ours \\
\midrule
NIQE$\downarrow$                     & 7.070          & 6.296                  & 6.638            & \textbf{6.171}   \\
MANIQA$\uparrow$                     & 0.6476         & 0.6460                 & 0.6505           & \textbf{0.6530}  \\
DISTS$\downarrow$                    & 0.2483         & 0.2402                 & 0.2518           & \textbf{0.2296}  \\
\bottomrule
\label{tab5}
\end{tabular}
}
\vspace{-8mm}
\end{table}
%
In addition, if we skip the redundancy analysis in the fine pruning stage and naively prune the same number of decoder blocks in the denoising UNet (``$\text{w/~Unfine}$'' for short), the computation is similar, but the restoration quality degrades, where the DISTS is 0.0274 worse (see comparisons of ``$\text{w/~Unfine}$'' and ``Ours'' in Table~\ref{tab4}),  which suggests that the redundancy analysis is important in the fine pruning stage and also demonstrates the effectiveness of the fine pruning stage.
Figure~\ref{fig6} shows the visual comparison on the DrealSR~\cite{DrealSR} dataset.
Compared to comparative methods, the results generated by our approach are more realistic and faithful, without artifacts on the windows.
\vspace{-1.5mm}

\subsection{Effect of the proposed PEAM}
\vspace{-0.5mm}
We provide the phase-exchange adapter module that uses the phase of the LQ image to guide the pruned diffusion model for better image restoration. 
To validate the effectiveness of the proposed PEAM, we compare it with three baselines. 
The first baseline method adds only the LQ features extracted by the adapter to the features in the pruned diffusion model (``w/ Add'', for short).
The second baseline method replaces only the phase information of the LQ features extracted by the adapter with the pruned diffusion model features (``w/ Phase change'' for short).
The third baseline method, from which our proposed method removes the fusion method, retains eleven stacked $3{\times}3$ convolutional layers (``w/ Fusion conv'' for short).
Comparisons of ``w/ Add'' and ``w/ Phase change'' in Table~\ref{tab5} show that directly using the phase information of the LQ image to guide the pruned diffusion model can generate better results that improve structure but slightly harm perception, where the NIQE and DISTS values are better $0.774$ and $0.0081$, respectively, but at the cost of a lower MANIQA value. 
In contrast, our proposed method exchanges features as dynamic guidance and adaptively fuses them with the original encoder features, achieving better restoration results, where the NIQE, DISTS, and MANIQA values are $0.125$, $0.0106$, and $0.0070$ better (see comparisons of ``w/ Phase change'' and ``Ours'' in Table~\ref{tab5}).
%

%
\begin{figure}[!t]
\footnotesize
\centering
\setlength{\tabcolsep}{2pt}
\begin{tabular}{ccc}
\includegraphics[width=0.32\linewidth,height=0.25\linewidth]{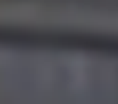} &
\includegraphics[width=0.32\linewidth,height=0.25\linewidth]{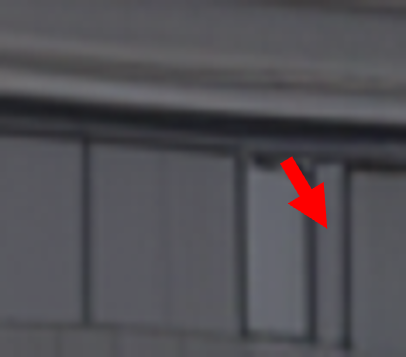} &
\includegraphics[width=0.32\linewidth,height=0.25\linewidth]{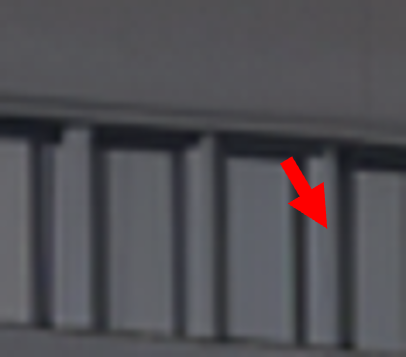} \\
(a) {\small LQ patch} &
(b) {\small  w/o PPA} &
(c) {\small $\text{w/o~Coarse}$} \\
\includegraphics[width=0.32\linewidth,height=0.25\linewidth]{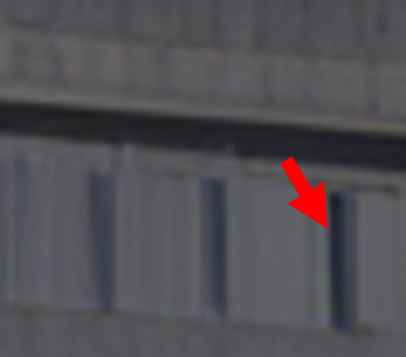} & 
\includegraphics[width=0.32\linewidth,height=0.25\linewidth]{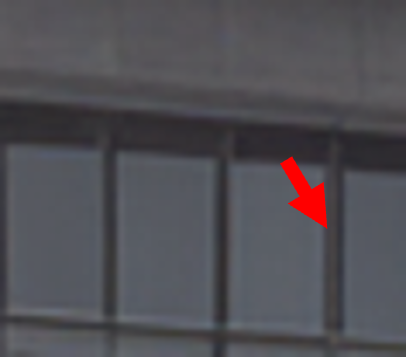} &
\includegraphics[width=0.32\linewidth,height=0.25\linewidth]{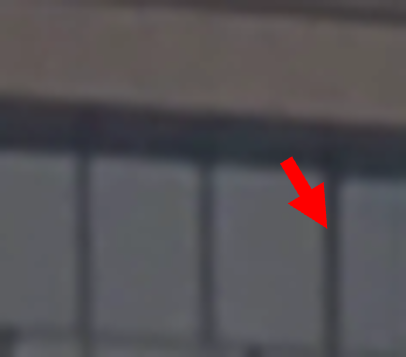} \\
(d) {\small $\text{w/ Unfine}$} & 
(e) {\small Ours}& 
(f) {\small GT patch} \\
\end{tabular}
\vspace{-4mm}
\caption{Effectiveness of the proposed PPA for PGP-DiffSR on image super-resolution.}
\label{fig6}
\vspace{-3mm}
\end{figure}
\begin{figure}[!t]
\footnotesize
\centering
\setlength{\tabcolsep}{2pt}
\begin{tabular}{ccc}
\includegraphics[width=0.32\linewidth,height=0.25\linewidth]{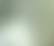} &
\includegraphics[width=0.32\linewidth,height=0.25\linewidth]{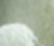} &
\includegraphics[width=0.32\linewidth,height=0.25\linewidth]{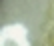} \\
(a) {\small LQ patch} &
(b) {\small w/ Add} &
(c) {\small w/ Phase change} \\
\includegraphics[width=0.32\linewidth,height=0.25\linewidth]{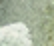} &
\includegraphics[width=0.32\linewidth,height=0.25\linewidth]{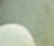} &
\includegraphics[width=0.32\linewidth,height=0.25\linewidth]{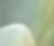} \\
(d) {\small w/ Fusion conv} & 
(e) {\small Ours}& 
(f) {\small GT patch}\\
\end{tabular}
\vspace{-4mm}
\caption{Effectiveness of the proposed PEAM for PGP-DiffSR on image super-resolution.}
\label{fig7}
\vspace{-7mm}
\end{figure}

%
In addition, the comparisons of ``w/ Fusion conv'' and ``Ours'' in Table~\ref{tab5} show that the results arise from the phase guidance itself rather than from the network capacity.
Figure~\ref{fig7} shows the visual comparison on the DrealSR~\cite{DrealSR} dataset. 
Compared to the comparative methods, the results generated by our approach are more realistic and faithful.
\vspace{-2.5mm}
\section{Conclusion}
\label{sec:conclusion}
\vspace{-1.5mm}
We have proposed a PGP-DiffSR that removes redundant information from diffusion models under the guidance of the phase information of inputs for efficient image super-resolution.
To remove redundant blocks in the diffusion model, we develop a simple yet effective PPA to remove redundant blocks in the encoder, bottleneck, and decoder modules of the denoising UNet.
Furthermore, we develop a PEAM to utilize the phase information from the LQ inputs and guide the pruned diffusion model for better restoration.
Extensive experiments on the benchmarks demonstrate that our method achieves competitive performance compared with state-of-the-art methods while substantially reducing computational and memory requirements.

{
    \small
    \bibliographystyle{ieeenat_fullname}
    \bibliography{main}
}

\end{document}